%% file: main.tex
\documentclass[conference]{IEEEtran}
\usepackage{times}
\usepackage{epsfig}
\usepackage{graphicx}
\usepackage{amsmath}
\usepackage{amssymb}
\usepackage{booktabs}
\usepackage{longtable}
\usepackage{circledsteps}
\usepackage[export]{adjustbox}
\usepackage{algorithm,algpseudocode}
\usepackage{subfig}
\usepackage{multirow}
\usepackage[pagebackref=true,breaklinks=true,colorlinks,bookmarks=false]{hyperref}
\usepackage[T1]{fontenc}
\usepackage[font=small,labelfont=bf,tableposition=top]{caption}
\usepackage{babel}
\usepackage[utf8]{inputenc}
\DeclareUnicodeCharacter{03B2}{$beta$}
\usepackage{footmisc}

\begin{document}

\title{Novelty-based Generalization Evaluation for Traffic Light Detection}
\author{
\IEEEauthorblockN{
Arvind Kumar Shekar\IEEEauthorrefmark{1},
Laureen Lake\IEEEauthorrefmark{1}, Liang Gou\IEEEauthorrefmark{2} and
Liu Ren\IEEEauthorrefmark{2}
}
\IEEEauthorblockA{\IEEEauthorrefmark{1}Robert Bosch GmbH
Stuttgart, Germany\\
\IEEEauthorrefmark{2}
Robert Bosch Research and Technology Center,
California, USA\\
\IEEEauthorrefmark{1}first.lastname@de.bosch.com,
\IEEEauthorrefmark{2}first.lastname@us.bosch.com}
}

\maketitle
\input{abstract}
\input{introduction}
\input{related_work}
\input{generalization_work}

\input{experiments}

\input{imple_det}
\input{conclusion}

{\small
\bibliographystyle{ieee_fullname}
\bibliography{bibli}
}

\end{document}

%% file: abstract.tex
\begin{abstract}
The advent of Convolutional Neural Networks (CNNs) has led to their application in several  domains. 
One noteworthy application is the perception system for autonomous driving that relies on the predictions from CNNs. 
Practitioners evaluate the generalization ability of such CNNs by calculating various metrics on an independent test dataset. 
A test dataset is often chosen based on only one precondition, i.e., its elements are not a part of the training data. 
Such a dataset may contain objects that are both similar and novel w.r.t. the training dataset. 
Nevertheless, existing works do not reckon the novelty of the test samples and treat them all equally for evaluating generalization. 
Such novelty-based evaluations are of significance to validate the fitness of a CNN in autonomous driving applications. 
Hence, we propose a CNN generalization scoring framework that considers novelty of objects in the test dataset. 
We begin with the representation learning technique to reduce the image data into a low-dimensional space. 
It is on this space we estimate the novelty of the test samples. 
Finally, we calculate the generalization score as a combination of the test data prediction performance and novelty. 
We perform an experimental study of the same for our traffic light detection application. 
In addition, we systematically visualize the results for an interpretable notion of novelty.
\end{abstract}

%% file: introduction.tex
\section{Introduction}\label{sec:intro} 
In recent times, machine learning models are increasingly deployed for several real-world applications \cite{ijcvarvind}. 
This has led to the development of several Convolutional Neural Network (CNN) architectures. 
It is a common practise to compare the CNNs based on their generalization ability. 
Generalization in computer vision and machine learning is defined as "\textit{the ability of image and video processing algorithms such as object detectors or classifiers to perform well not only on the dataset they are trained on, but also on novel data} \cite{general_defn}." 
Hence, the generalization of the developed networks are based on their performance on a test dataset that is independent of the training data \cite{he2019control}. 
However, such an independent dataset is often chosen randomly. 
A random selection does not ensure that the test subset has novel samples. 
In AD application, we acquire data from several video streams at regular intervals \cite{ijcvarvind}. 
In principle every image in the stream is independent from the training dataset. 
However, for generalization evaluation it is necessary to gauge the network's performance on novel images that the network did not see during the training. 
Hence, we introduce a second criterion, i.e., novelty of the objects in the image, to weigh the network's performance. 

Given a training dataset, a test sample is deemed novel when they are anomalous w.r.t. it \cite{noveltydiscreteseq}. 
Such anomalous samples do not conform with the general properties of the majority of the training samples and are present in the low density regions of the training dataset. 
Therefore, the anomalous samples are not to be confused with mislabelled data. 
The terms outlier and novelty detection are synonymous to each other because they both detect samples that are anomalous. 
In fact, few works categorize both outlier and novelty detection as anomaly detection \cite{tan2011fast} and outlier detection methods have been applied for novelty detection in several cases \cite{noveltyocsvm,noveltydiscreteseq,lee2015novelty,markou2003novelty,reviewofnovelty}. 
The underlying difference between them lies in the step that follows detection.
The former aims at eliminating the detected anomalies, whereas the latter aims to identify new samples (w.r.t. a normal dataset) from the data \cite{mendelson2020online}. 
Although the occurrence of novel samples may be infrequent in comparison to regular samples, evaluating the CNNs performance on them is not a trivial task \cite{reviewofnovelty}. 
Such evaluations are important in AD applications to understand the limitations of a CNN prior to deployment. 
Nevertheless, the major challenge for novelty-based generalization is the magnitude of the AD datasets, i.e., they consist of several thousands of images with multiple objects. 
Evaluating the novelty of each image in the test dataset w.r.t. all training images based on manually defined properties, e.g., brightness, blurriness, is tedious \cite{7004298,ijcvarvind}. 

We address this challenge by using the representation learning method, i.e., Variational Auto-encoders (VAE) \cite{betavae}, to extract the key features of the dataset, i.e., latent dimensions. 
Given a dataset of size $N \times H \times W \times C \times$, where $N$ is the number of images, $H$ is its height, $W$ is its width and $C$ is the number of channels, the VAE reduces this to a $N\times d$ space, where $d$ is the number of latent dimensions. 
It is on this low-dimensional latent representations we perform our novelty assessment. 
By this way, we avoid the need to perform and one-to-one comparison between the train and test data. 
Finally, we score the generalization of a CNN by rewarding its ability to perform well on the high novel samples. 
Few existing works apply representation learning, e.g., Auto Encoders (AE), to identify novel samples based on the reconstruction errors \cite{chen2018autoencoder}. 
However, over several iterations AE reduces the reconstruction error of the novel samples as well and may not be directly applicable for novelty estimation \cite{robustae}. Hence, we use them only as a dimensionality reduction framework. 
We prefer a VAE instead of an AE because the representations of an AE are not continuous. 
Whereas in case of a VAE, the latent representation being mean value of a distribution can be varied to understand the physical meaning of that dimension. 
This property is essential for human interpretation of novelty \cite{VATLD}. 
Existing works target on enhancing the CNN generalization during the training process by regulating several factors such as number of training iterations, learning rate, batch size and regularization \cite{goodfellow2016deep,he2019control,zheng2018improvement}. 
But to the best of our knowledge, there are no existing KPIs (Key Performance Indicator) that quantify the generalization ability of a fully trained CNN. 

In this work, we propose a two step framework to quantify the generalization ability of a CNN.
In the first step we reduce the dimensionality of the training and test dataset using a VAE. 
In the second step we estimate the novelty of a test sample by evaluating its likelihood that it belongs to the training data. 
Finally, the prediction performance of a CNN on a given test sample is weighted by the novelty of the test sample to obtain a single generalization score. 
Moreover to ensure interpretability of the novelty scorer, we visualize its results and present a more human-understandable notion of novelty. 
The primary contributions of this work are, 
\begin{itemize}
\item A generalization scoring measure based on representation learning and novelty detection algorithm. 
\item Performance comparison of various novelty detection algorithms from different paradigms.
\item Generalization evaluation experiments on different CNN architectures for traffic light detection problem in the AD domain. 
\item Interpretation of novelty using the latent representations learnt by the VAE.
\end{itemize}


%% file: related_work.tex
\section{Related work} \label{sec:related_work} 
Novelty detection literature is broadly grouped into classification, e.g., One-Class Support Vector Machines (OCSVM), Nearest Neighbor, e.g., Local outlier factor (LOF), clustering, e.g., k-Nearest Neighbors (kNN) and statistical based approaches, e.g., Histogram-based outlier score \cite{lof,goldstein2012histogram,masud2010classification,reviewofnovelty,ocsvm}. The works of \cite{markou2003novelty,reviewofnovelty} provide a brief summary about the novelty scoring taxonomies and algorithms. 
As the scoring algorithms suffer from the curse-of-dimensionality, it is often preceded by a feature extraction or dimensionality reduction step \cite{ding2014experimental}. 

As a classification-based approach, the work of \cite{noveltyocsvm} applies OCSVM for monitoring novel patterns in sensor data. 
They firstly extract statistical features from various sensors, which are then passed on to OCSVM for novelty scoring.
Instead of feature extraction, the work of \cite{lian2012feature} performs dimensionality reduction using Principal Component Analysis \cite{goodfellow2016deep} before using OCSVM. 
Likewise, the work of \cite{lee2015novelty} extracts the word usage patterns from documents to score the novelty of patents using LOF \cite{lof}.  
The work of \cite{masud2010classification} uses kNN novelty detection on data streams to learn the boundary of the training data. 
The test sample novelty is based on its distance from the trained boundary. 
Histogram based outlier score (HBOS) is a simple combination of uni-variate methods applied to multiple dimensions \cite{goldstein2012histogram}. 
Although, they do not account for dependence between variables, its efficiency on large datasets is an advantage. 
Similarly, Lightweight Online Detector of Anomalies (LODA) \cite{pevny2016loda} uses an ensemble of one-dimensional histograms for novelty detection. 
The work of \cite{mendelson2020online} applies it for novelty detection in data streams. 
The work of \cite{noveltydiscreteseq} uses LSTM-AE to reduce the discrete sequences into latent representations. 
During testing, the test data samples are reduced into the latent representations and fed as input to the Isolation Forest \cite{liu2008isolation}. 
The Isolation Forest is advantageous as they are unsupervised and isolate the novel data points without the need of normal data. 

Unlike the straightforward AE reconstruction error based novelty scorer \cite{chen2018autoencoder}, the work of \cite{robustae} introduces likelihood of a test sample in addition. 
Similarly, the work of \cite{angiulli2010outlier} applies a likelihood ratio test for evaluating the novelty of the video sequences based on predefined normal data. 
The q-space novelty detection method \cite{vasilev2020q} proposes several novelty scoring methods using the VAE latent feature and original space. 
One of them quantifies novelty by estimating the likelihood of a test sample belonging to the modeled latent space distribution of the normal data. 
In contrast to aforementioned methods, we apply the novelty detection on image datasets in the context of CNN generalization evaluation. 
Moreover, we experimentally present a systematic performance comparison of various novelty scoring approaches. 
Secondly, we do not use the AE reconstruction error as a novelty score and rather use VAE only for dimensionality reduction. 
Although, we estimate novelty based on likelihood of the test sample as done in the work of \cite{robustae}, we additionally visualize novelty on the latent and the image space. 
That is, we provide an interpretable notion of novelty rather than mere reporting of scores. 

%% file: generalization_work.tex
\section{Our framework}\label{sec:approach} 
\begin{figure}[h]
\centering
\includegraphics[width=0.48\textwidth]{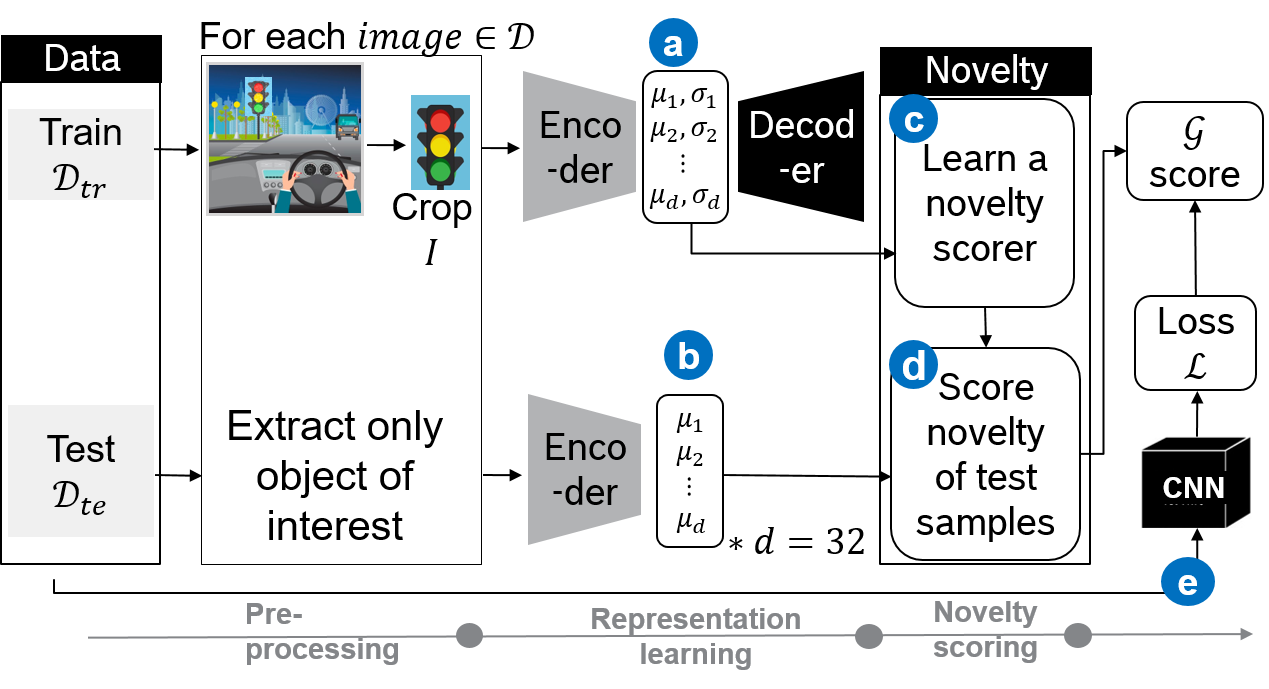}
\caption{Our generalization framework}
\label{fig:novelty_fw}
\end{figure}
Our generalization evaluation framework involves three phases, viz., pre-processing, representation learning and novelty scoring. 
We explain each of them using traffic light (TL) localization CNN from our Autonomous Driving (AD) application. 
Our framework aims to score the generalization ability of the trained traffic light detectors. 
Given a traffic scenario image for the task of TL detection, it contains several irrelevant information or objects and evaluating their pixel-wise novelty is inefficient and less insightful. 
Hence, similar to the work of \cite{VATLD}, we start with the pre-processing phase where the objects of interest, i.e., traffic lights, are cropped from the train and test datasets. 
On one hand, the pre-processing phase reduced the size of the images by targeting the object-of-interest $I$. 
On the other hand, pre-processing phase will increase the number of images because each traffic scenario may have more than one TL object and each of them is extracted as a separate image. 
For example, the Bosch Small Traffic light dataset and the subset of DriveU dataset we use has $\approx$ 10k and 50k traffic lights in the training data \cite{bstld, dtld}. 
Hence, we pass the TL crops to the representation learning phase to train a Variational Auto-Encoder (VAE) (c.f. Figure \ref{fig:novelty_fw} {\Circled[inner color=white, outer color=blue, fill color=blue]{\textbf{a}}). 

The VAE comprises of an encoder, a $d$-dimensional bottleneck layer and a decoder which mirrors the encoder. 
Using the training data, its objective is to learn a $d$-dimensional latent distributions by optimizing a set of parameters $\phi$ based on the KL-divergence and reconstruction loss. 
Each dimension is defined by a mean $\mu$ and standard deviation $\sigma$ . 
Hence, a fully trained encoder $e_{\phi}$ is capable of mapping a TL crop to $d$ mean and standard deviation values. 
For deeper understanding of VAE, readers may refer to the original work \cite{betavae}. 
We use the mean value as a low-dimensional representation of the traffic light crop $I$, $e_\phi:I \mapsto \mu \mid |\mu|=d$. 
These representations hold the key properties of the RGB images in a compressed form. 
For example, we project the $d$-dimensional $\mu$ vector ($d=32$) into a 2-dimensional space using UMAP \cite{mcinnes2018umap} and DriveU data in Figure \ref{fig:umap_tls}. 
The first observation is that the $\mu$ values cluster themselves based on the color of the lights. 
Although this work focuses on TL detection and not classification, with Figure \ref{fig:umap_tls} we intend to emphasize that the lights of same color may have different properties. 
That is, each color cluster has both dense and sparse regions. 
Given a new test image, our target is to compress it to the $d$-dimensional $\mu$ vector and score its novelty w.r.t. the training data. 
For the sake of completeness, we trained a simple AE over the same dataset. However, we did not observe such clustering of the samples based on colors. 
Moreover, its latent dimensions were challenging to interpret. 
Whereas, to understand a latent dimension of a VAE we traversed through the learnt distribution to observe the change in the reconstructed image.
\begin{figure}[h]
	\centering
	\includegraphics[width=0.45 \textwidth, keepaspectratio]{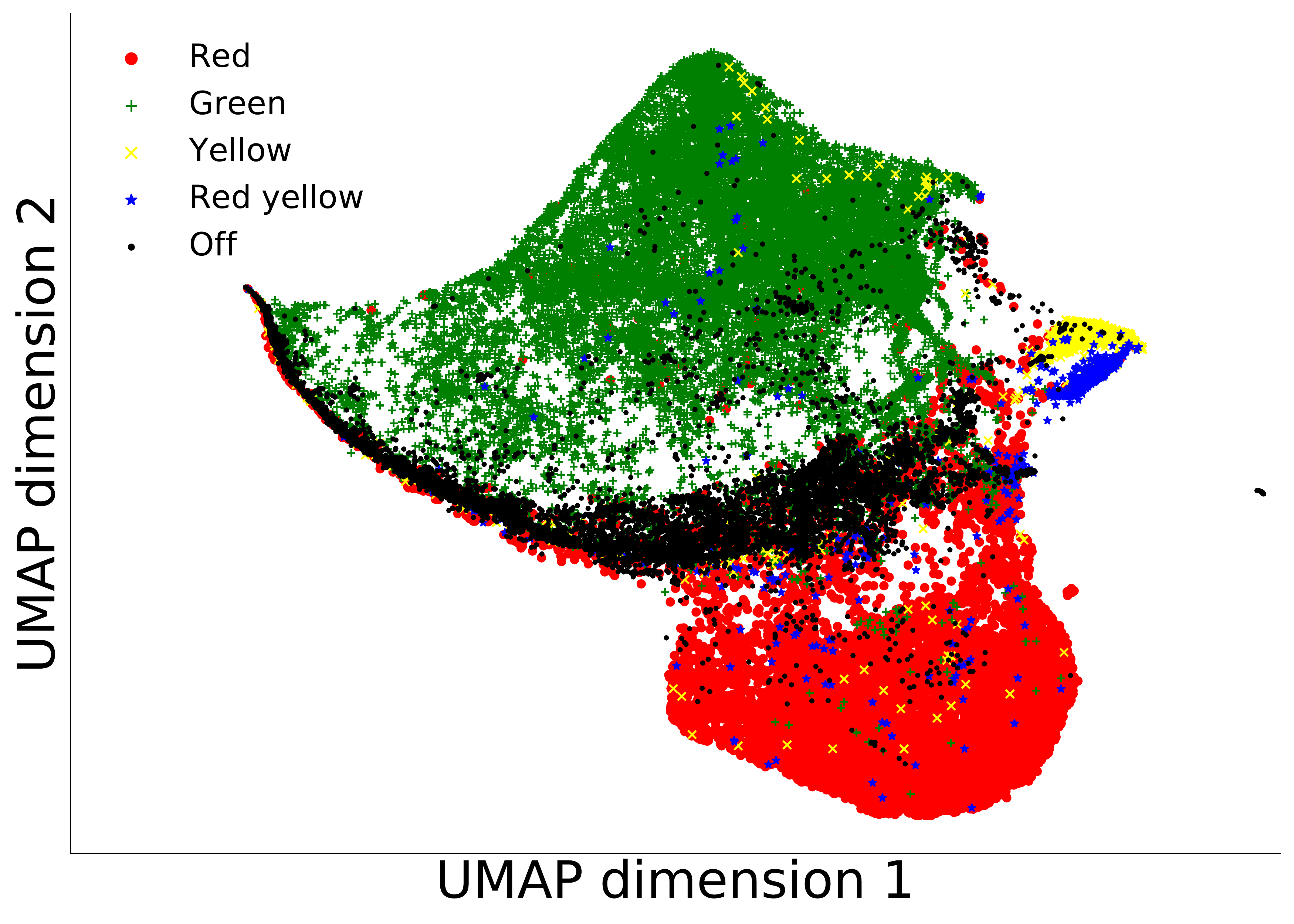}
	\caption{2-dimensional representation of the VAE $\mu$ vector using DriveU dataset}
	\label{fig:umap_tls}
	\vspace{-1.5em}
\end{figure}

The encoder $e_{\phi}$ is capable of transforming the train and test images into a low-dimensional mean vectors (c.f. Figure \ref{fig:novelty_fw} {\Circled[inner color=white, outer color=blue, fill color=blue]{\textbf{a}}} and {\Circled[inner color=white, outer color=blue, fill color=blue]{\textbf{b}}}). 
Given $N$ training and $\bar{N}$ test images, it is transformed into $Z\in\mathbb{R}^{N,d}$ and $\bar{Z}\in\mathbb{R}^{\bar{N},d}$ respectively. 
We train a novelty scorer using the $Z$ matrix and score the novelty of each test sample in $\bar{Z}$ (c.f. Figure \ref{fig:novelty_fw} {\Circled[inner color=white, outer color=blue, fill color=blue]{\textbf{c} and \textbf{d}} respectively). 
After calculating the novelty score for each TL crop, the next step is to pass the full images (un-cropped) to the TL detector CNN (c.f. Figure \ref{fig:novelty_fw} {\Circled[inner color=white, outer color=blue, fill color=blue]{\textbf{e}}). 
Just as the novelty scorer assigned a score for each TL crop $nov\_sc:(I,e_{\phi})\mapsto novelty\in\mathbb{R}$, each CNN prediction is assigned with a confidence score $s\in \mathbb{R}$. 
A prediction is either a true positive, i.e., the predicted box $\hat{y}$ overlaps the ground truth box $y$ with an IoU greater than 0.5, or a false positive if otherwise. 
In the case of a false negative, we assign that object $I$ a confidence of 0. 
Since the false positives do not have a novelty score, we do not use it for the generalization scoring. 
As a network performance measure, we calculate the loss $\mathcal{L}:(y,\hat{y},s)\mathbb{R}$ for each detection. 

Overall, each TL crop now has a novelty score and a performance score in the form of a loss value. 
Our generalization score rewards a low loss for highly novel objects by calculating a novelty weighted average of the loss $\mathcal{G}= \frac{\sum_{i=1}^{N}novelty_i \cdot  (1-\mathcal{L}(y_i,\hat{y}_i,s_i))}{\sum_{i=1}^{N}novelty_i}.$ 
That is, a loss value of a highly novel sample is weighted higher than that of less novel sample. 
However, it is worth mentioning that we do no ignore the less novel samples. 
Rather, we only weigh them lower in comparison to the highly novel samples. 
Therefore, our generalization score is a combination of both high and low novel samples but aggregated with different weights.

%% file: experiments.tex
\section{Experimental evaluations}\label{sec:experimental_eval} 
As mentioned in Section \ref{sec:approach} we use two well-known TL datasets, i.e., BSTLD and DriveU for the experimental evaluations. 
We instantiate the $\mathcal{L}$ with Mean Absolute Error (MAE) because it has fixed upper and lower-bounds $[0,1]$. 
Our evaluation targets to, (a) compare various novelty scoring algorithms that fits our application, (b) evaluate usefulness of the proposed generalization score $\mathcal{G}$ and (c) interpret the notion of novelty for the datasets. 
We extracted the TL crops from both datasets to train the VAE for 750 epochs, with a batch size of 64, a learning rate of 0.0001, $d$=32 and $\beta=0.1$. 


\subsection{Comparative study}\label{sec:benchmark_study}
The novelty scorer for the TL objects (crops) is an important part of our generalization framework (c.f. Figure \ref{fig:novelty_fw} in step {\Circled[inner color=white, outer color=blue, fill color=blue]{\textbf{c}} and {\Circled[inner color=white, outer color=blue, fill color=blue]{\textbf{d}}). 
In this section we perform a comparative evaluation of existing methods to identify a scorer that best fits our application. 
For this, we select seven novelty scoring algorithms from various paradigms (c.f. Section \ref{sec:related_work}). 
As we do not have the ground truth novelty scores for the study, we systematically engineer the datasets for our experiment. 
First we choose a contamination color, e.g., green TLs, which is removed from the train dataset. 
We engineer the test dataset such that it contains 10\% of the contamination color and 90\% of the other colors. 
The contamination color samples are novel w.r.t. training data in this context because all contamination color samples were intentionally removed from the training dataset. 
We aim to compare the ability of different novelty scoring algorithms to identify these samples in the test dataset. 
We use the implementation from PyOD and scikit-learn in our evaluation \cite{zhao2019pyod} and the area under ROC curve as a quality metric for our evaluation. 
As a randomness component is involved in our experiment, we perform the contamination sampling three times and show the standard deviation and average ROC.

In Figure \ref{fig:benchmark}, we show the results on BSTLD \cite{bstld} and DriveU \cite{dtld} dataset with green and red as contamination colors respectively. 
Amidst the chosen scorers, LOF was the best performing (largest ROC score), followed by the KDE. 
However, with increase in the number of contamination (novel) samples, we observed that number false positives (normal samples identified as novel) increased more for LOF in comparison to KDE (see supplementary material \footref{note1}). 
For this reason, we prefer to use KDE as the novelty scorer for our further experiments. 
The density score of KDE is inversely proportional to novelty and it may have negative values. 
We normalise the density scores such that they are proportional to novelty and non-negative values.\footnote{Supplementary material: \url{https://figshare.com/s/2613fa8fddad96895a6f}\label{note1}}. 
\begin{figure}
\centering
\includegraphics[trim={0 0 1cm 1cm},clip, width=0.48\textwidth, keepaspectratio]{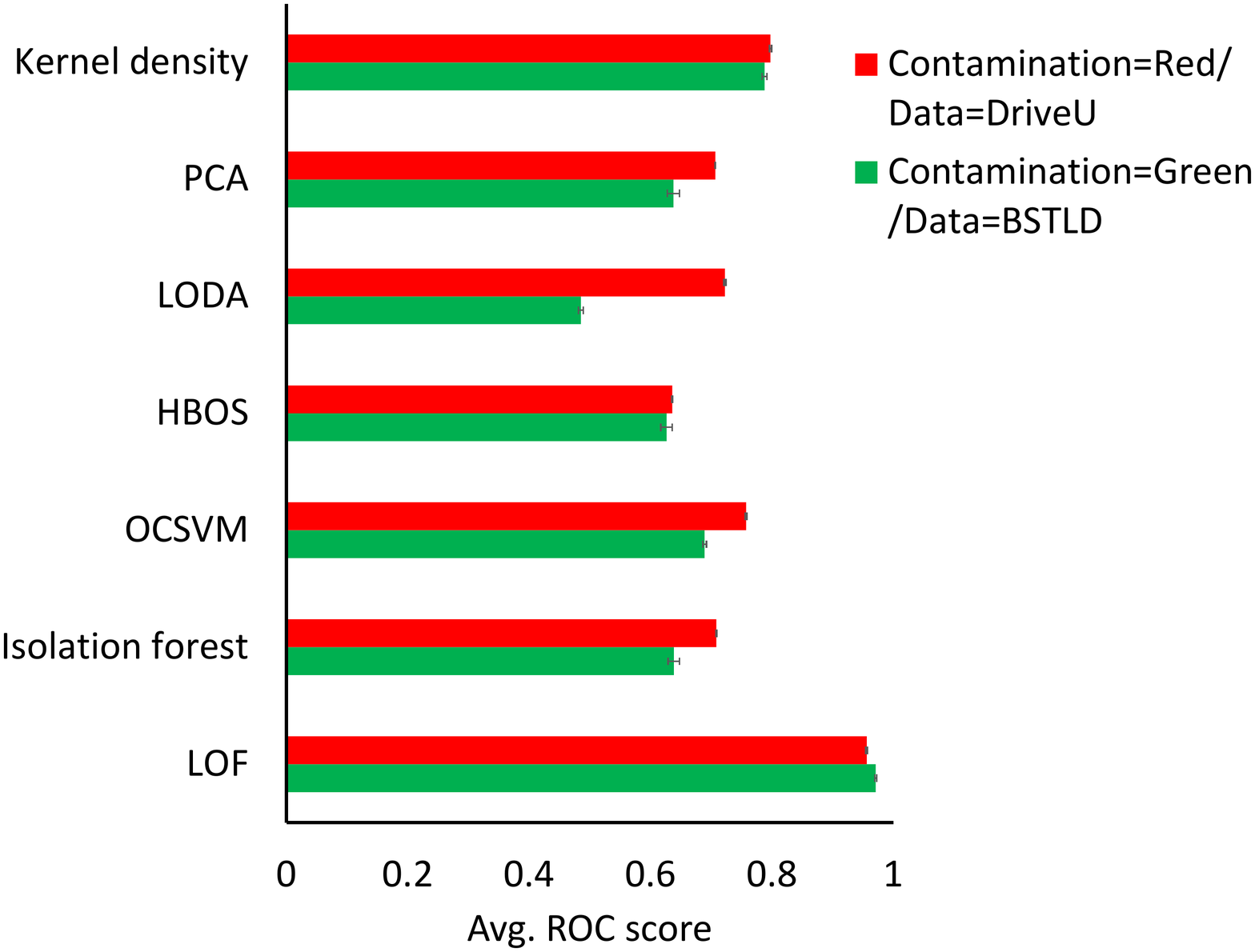}
\vspace{-0.5em}
\caption{Novelty algorithms comparative study}
\label{fig:benchmark}
\vspace{-2em}
\end{figure}


\subsection{Generalization comparison for TL detector}\label{sec:generalisation_tl_comparison}
\begin{figure}[!htp]
    \vspace{-2em}
	\centering
	\subfloat[Effect of dropout regularization using SSD (inception) and BSTLD dataset]{\includegraphics[width=0.4 \textwidth, keepaspectratio]{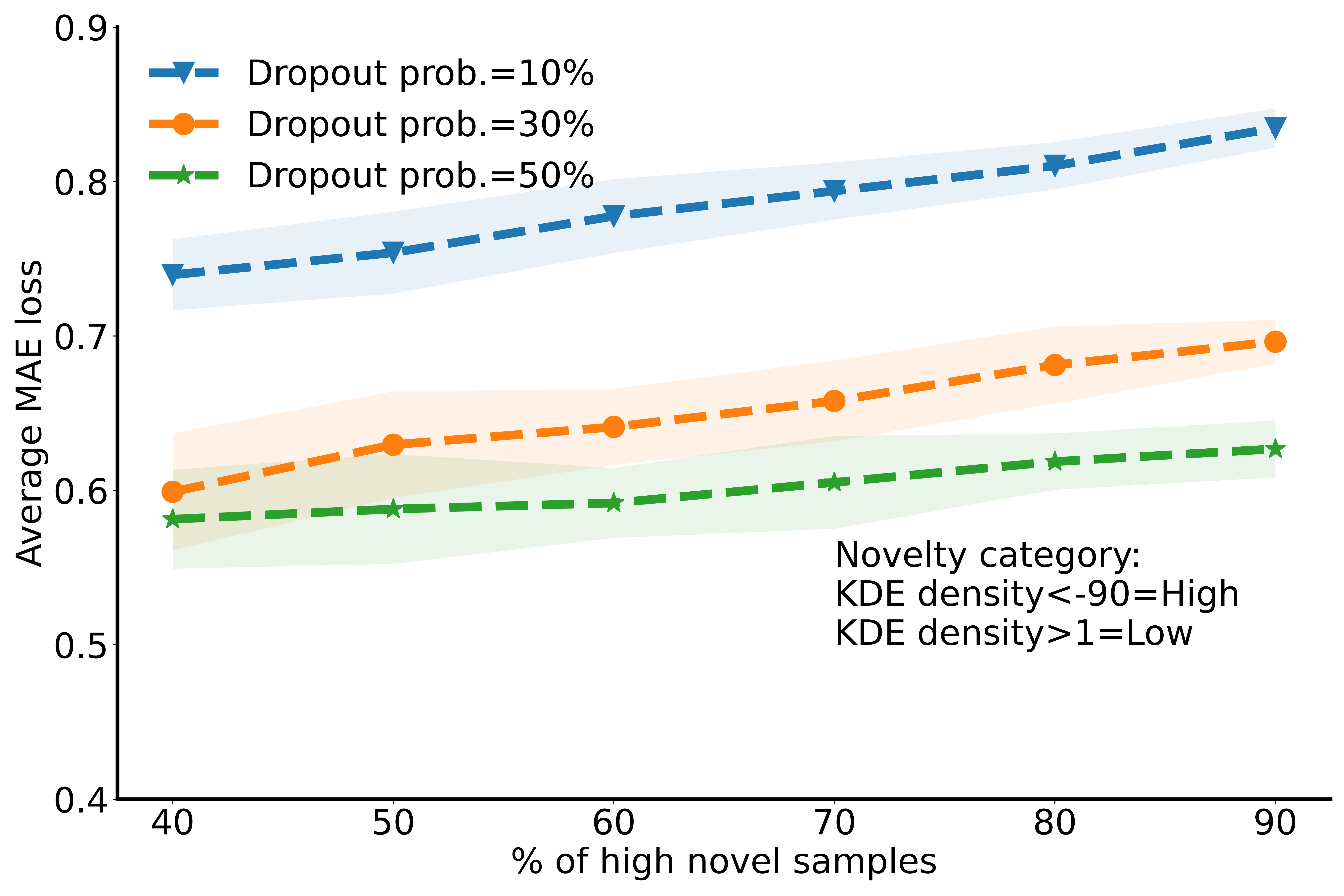}\label{fig:dropout_nov}}\\
	\subfloat[Effect of training iterations using BSTLD dataset]{\includegraphics[width=0.4\textwidth, keepaspectratio]{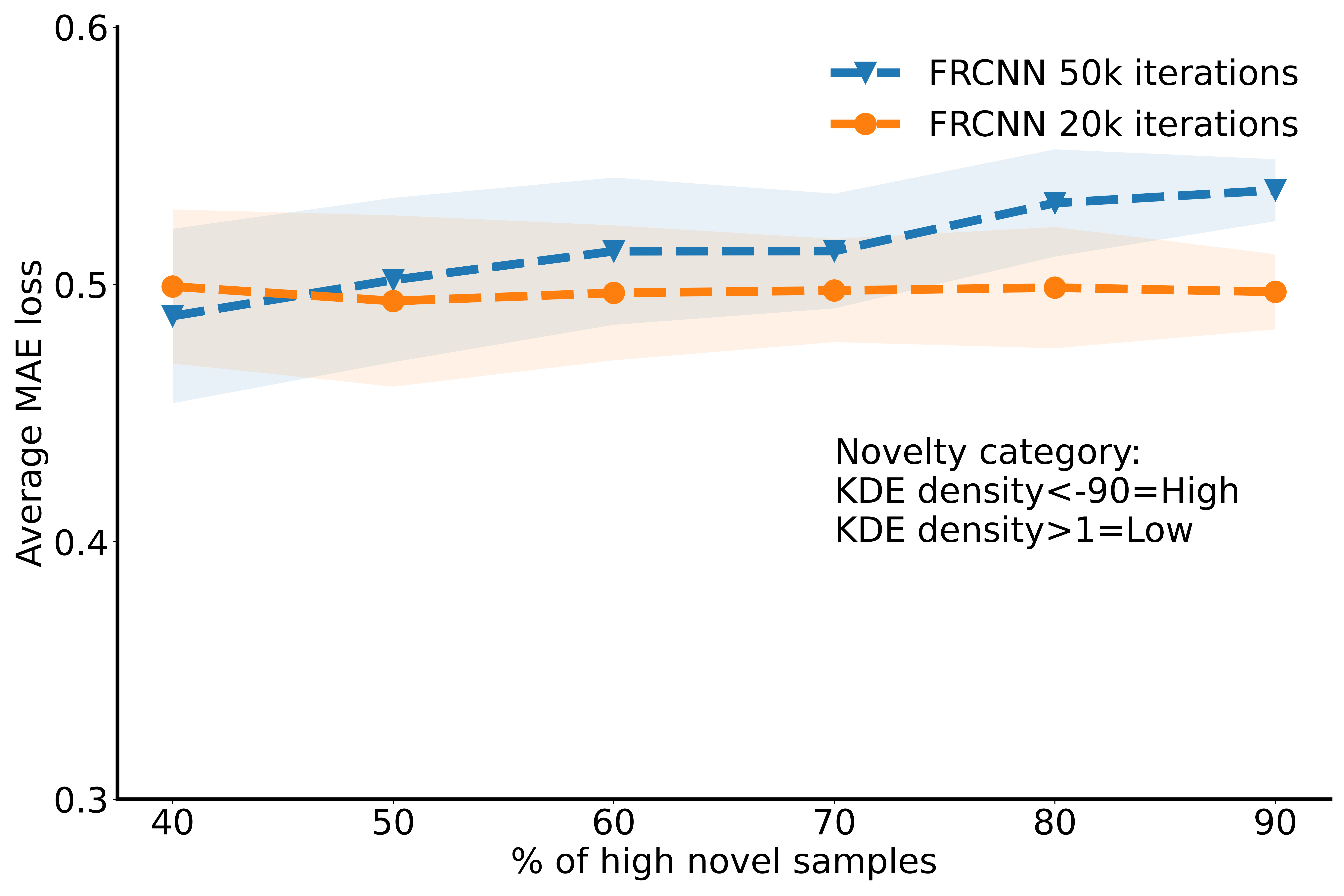}\label{fig:frcnn_iterations}} \\
	\subfloat[Effect of batch size using BSTLD dataset]{\includegraphics[width=0.4 \textwidth, keepaspectratio]{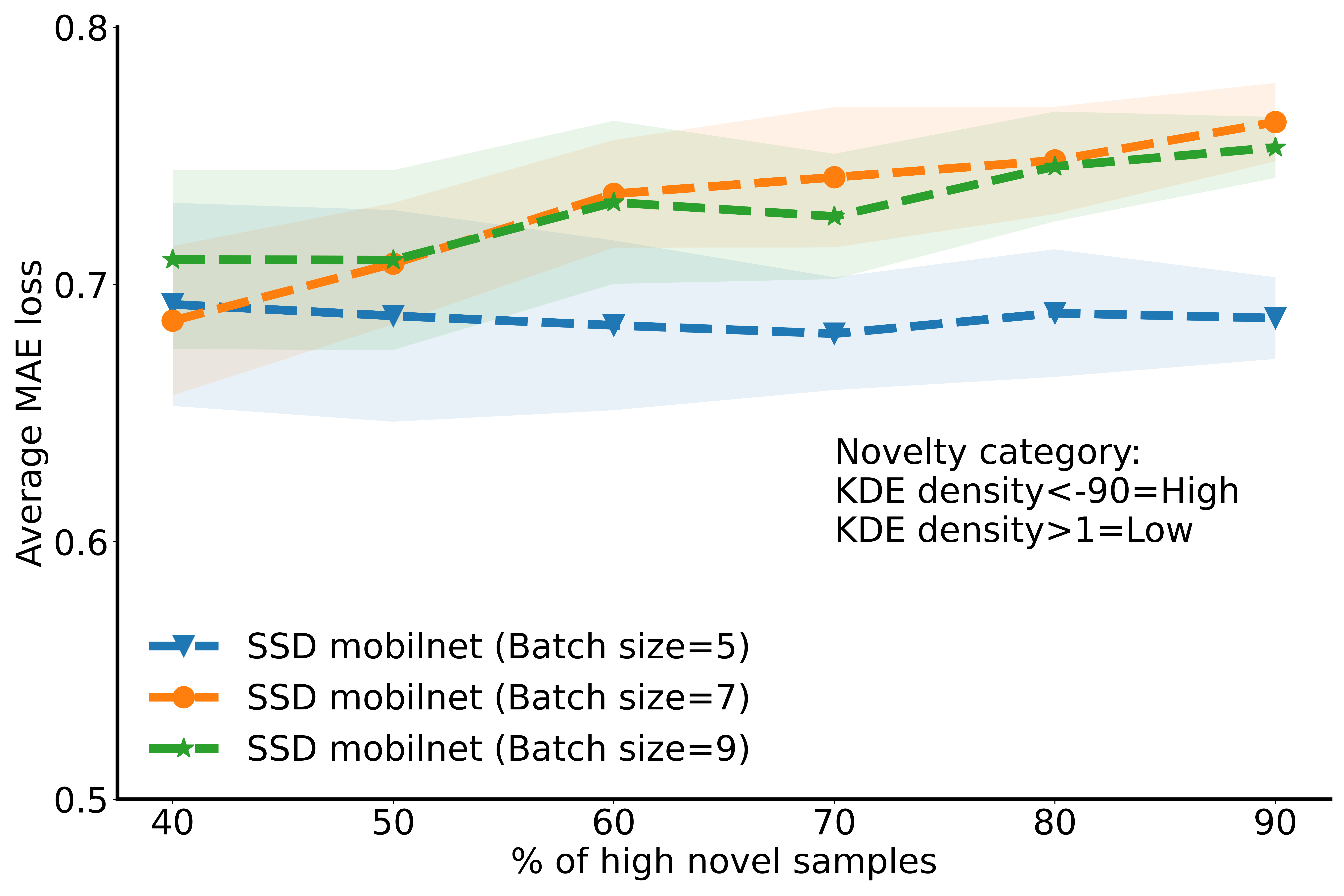}\label{fig:ssd_mobi_bs_diff}}\\
	\subfloat[Effect of learning rate using DriveU dataset and SSD inception v2]{\includegraphics[width=0.4 \textwidth, keepaspectratio]{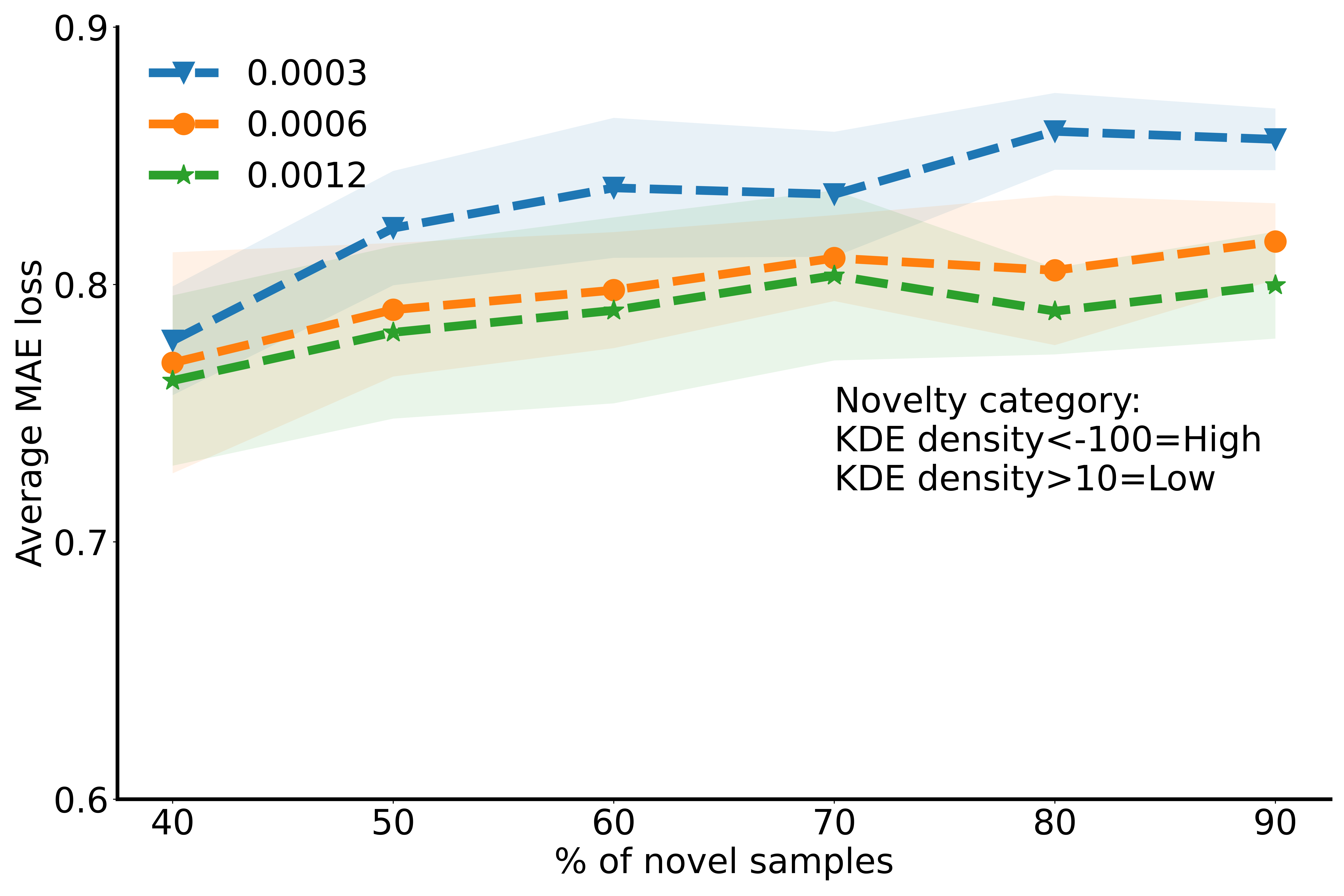}\label{fig:driveu_lrr_ssd}}
	\caption{Effect of various factors on the loss function}
	\label{fig:nov_vs_mae}
	\vspace{-2em}
\end{figure}
In this section we show the influence of the novelty scores on CNN generalization ability. 
However, we do not have a ground truth generalization score for benchmarking. 
For this reason, we use various factors from the existing works that are known to control the network's generalization ability, e.g., dropout rate, number of training iterations, batch size and learning rate. 
As the object-under-test we chose three CNN architectures, SSD and FRCNN with inception V2 backbone and SSD with mobilnet backbone. 
We follow the same evaluation method as in Section \ref{sec:benchmark_study}, except that we define the contamination samples based on the novelty and not TL colors. 
After scoring each TL object with a novelty score using KDE, we bin the test data into high, medium and low novel samples. 
Like outliers, novel samples are also infrequent in a dataset. 
To avoid the results being dominated by low novel samples, we sample 100 images from each novelty bin. 

In Figure \ref{fig:dropout_nov} we evaluate the contribution of dropout regularization to generalize on novel samples. 
In general, with increasing dropout rate we observe that the test loss of the SSDs were smaller, i.e., regularization positively contributes for the generalization. 
Specifically, the dropout rates of 30\% and 50\% starts almost at the same loss values when the data is dominated by low novel samples. 
However, with increasing high novel samples the SSD with 50\% drop out rate has lower loss in comparison to the other, i.e., it generalizes better. 

To further emphasize the importance of novelty we present experiments that exhibited similar trend, i.e., having similar loss on the low novel samples but increasing loss over increase in novelty.  
For example, in Figure \ref{fig:frcnn_iterations} we observe that the FRCNN detector with 50k and 20k iterations has a similar loss when the test dataset predominantly has less novel samples. 
However, the loss of FRCNN trained with 50k iterations increased steeply with increasing novelty. 
One possible explanation is that the model was over-fitting the train data in comparison to the FRCNN with 20k iterations. 

\begin{figure}
\centering
\includegraphics[width=0.45\textwidth]{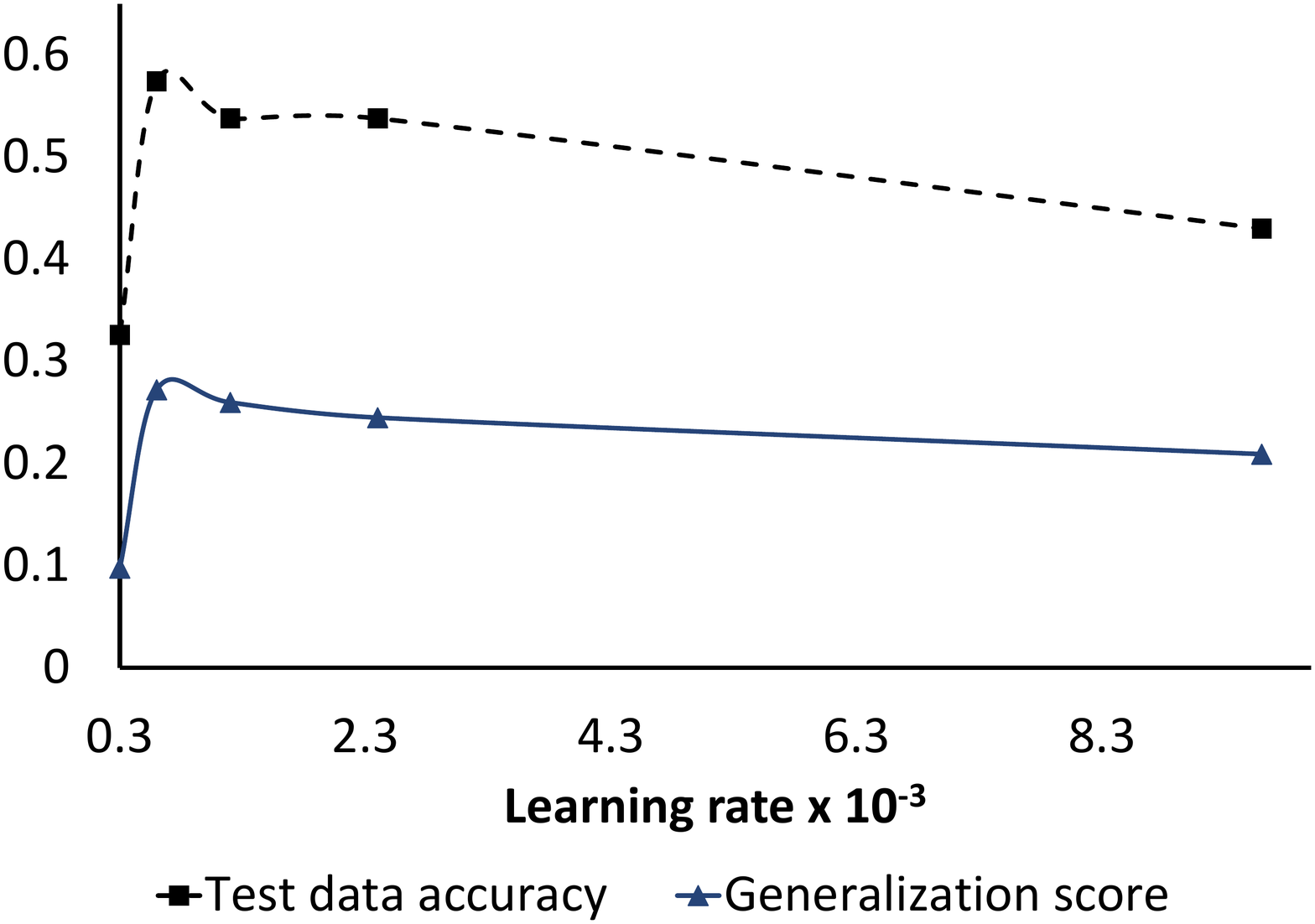}
\caption{Test data accuracy vs. our generalization score (all samples in BSTLD)}
\label{fig:generalization_benchmark_ssd_bstld_lrr_10k}
\end{figure}
\begin{table}
\centering
\resizebox{0.3\textwidth}{!}{
\begin{tabular}{@{}cccc@{}}
\toprule
Dataset & Network & Accuracy & $\mathcal{G}$ \\ \midrule
\multirow{2}{*}{BSTLD} & SSD & \textbf{0.65} & 0.23  \\ 
 & FRCNN & 0.6 & \textbf{0.33}  \\ \hline 
\multirow{2}{*}{DriveU} & SSD & 0.25 & 0.13\\ 
 & FRCNN & \textbf{0.5} & \textbf{0.44}  \\ \bottomrule 
\end{tabular}
}
\vspace{0.5em}
\caption{Test data accuracy vs. our generalization score (subset of BSTLD data with equal number of low/medium/high samples)}
\label{tab:generalization}
\vspace{-2em}
\end{table}
Batch size and network's generalization ability are negatively correlated, whereas learning rate and generalization ability are positively correlated \cite{he2019control}. 
From Figure \ref{fig:ssd_mobi_bs_diff} we infer the same. 
For instance, the SSD mobilnet of large batch size (7 and 9) shows a steep increase in the CNN loss as the number of novel samples increases. 
On contrary, the SSD with batch size 5 has a more stable or constant loss. 
However, this was not true for all networks, e.g., SSD inception with a batch size of 1 performed worse than batch size 5. 
On the other hand, from Figure \ref{fig:driveu_lrr_ssd} we observe that the SSD with learning rate 0.0006 and 0.0012 performed similar with increasing novelty. 
However, the SSD with the learning rate of 0.0003 shows comparatively larger loss with increasing novelty. Hence, the increase in learning rate does improve generalization, i.e., lower loss. 
However, setting the learning rate to extreme values will negatively affect the learning process. 
Although we used manual thresholds to bin the samples into high, medium and low novel samples (c.f. Figure \ref{fig:nov_vs_mae}), our generalization score does not require it. 
The novelty scores are used directly as weights for the $\mathcal{G}$ score computation. 

Conventionally the generalization of a CNN is evaluated based on test data accuracy. 
Hence we compare our generalization score $\mathcal{G}$ alongside test data accuracy. 
For this, we trained the SSDs at different learning rates on the two datasets separately. 
As we performed the learning rate analysis with loss values on DriveU dataset in Figure \ref{fig:driveu_lrr_ssd}, we show the results on BSTLD in Figure \ref{fig:generalization_benchmark_ssd_bstld_lrr_10k}. 
From the first look we observe that our generalization score follows the test data accuracy well. 
However, we demonstrate that our scoring is more insightful by including the novelty of the test objects with an additional experiment. 
As mentioned in Section \ref{sec:approach} test data accuracy is computed based on IoU between ground truth and predicted boxes. 
Using this as a generalization scorer we infer that the SSD generalizes better on BSTLD dataset (c.f. Table \ref{tab:generalization}). 
However, in general we observed that the loss values of FRCNN detector are smaller in comparison to the SSD.  
Especially on high novel samples the average loss values of SSD was 0.59, whereas that of FRCNN was 0.46. 
Our $\mathcal{G}$ score rewards this property to score FRCNN better than the SSD. 

\subsection{Interpretation of novelty}\label{sec:interpretation}
In the previous sections, we use the word novelty to describe objects that are anomalous w.r.t. the training dataset. 
However, in practical applications the user needs to interpret novelty to further enrich the dataset or improve the CNN architecture. 
Therefore, in this section we aim to understand the latent dimensions because KDE scores the novelty based on these dimensions. 
Instead of visualizing all the $d$-dimensions, we used mutual information \cite{VATLD} to select dimensions that best classifies the samples into high and low novelty categories. 
Therefore, we analyse only the dimensions that strongly influence the novelty of objects. 
In Figure \ref{fig:bstld_rel_traversals_bstld} and \ref{fig:rel_traversals_driveu} we visualize the traversal (low to high values) of the influential dimensions on BSTLD and DriveU datasets respectively. 
We understand that the dimensions represent various human-interpretable properties of the image like brightness, color shade, inlay or the bulb size. 

\begin{figure}
\vspace{-1em}
	\centering
	\subfloat[$z_{9}$ controls the size of the traffic light bulb]{\includegraphics[width=0.47\textwidth]{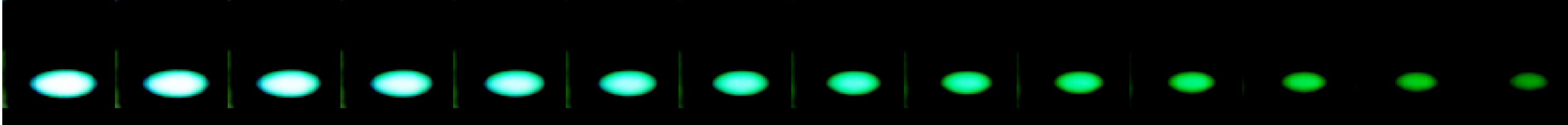}\label{fig:bstld_green_traversal_d9}} \vspace{-1em} \\
	\subfloat[$z_2$ controls the background brightness]{\includegraphics[width=0.47\textwidth]{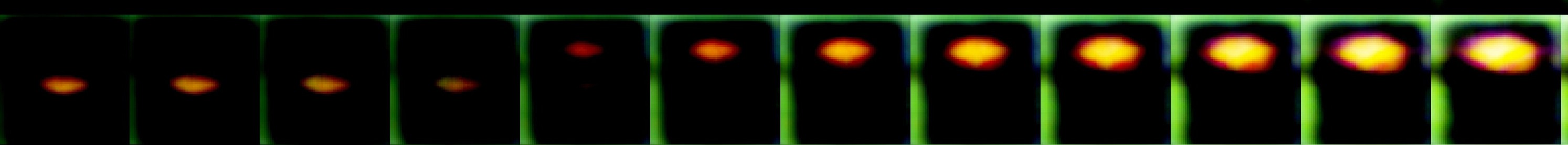}\label{fig:bstld_red_traversal_d2}} \vspace{-1em}\\
	\subfloat[$z_{19}$ controls the traffic light inlay]{\includegraphics[width=0.47\textwidth]{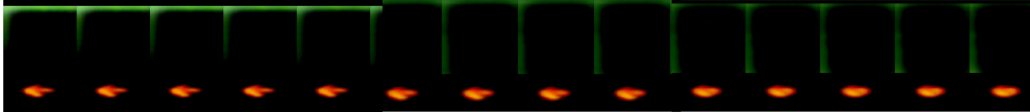}\label{fig:bstld_red_arrow_19}} 
	\caption{Traversal of VAE latent dimensions trained on BSTLD dataset}
	\label{fig:bstld_rel_traversals_bstld}
	\vspace{-1em}
\end{figure}
\begin{figure}
    \centering
	\subfloat[$z_{10}$ controls the white spot on the green traffic light]{\includegraphics[width=0.47\textwidth]{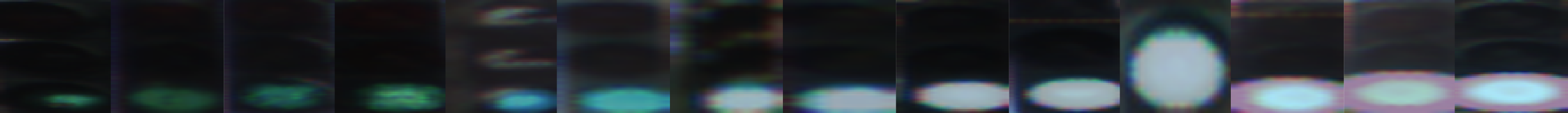}\label{fig:driveu_green10_traversal_whitespot}} \vspace{-1em} \\
	\subfloat[$z_{5}$ controls the shades of yellow color light]{\includegraphics[width=0.47\textwidth]{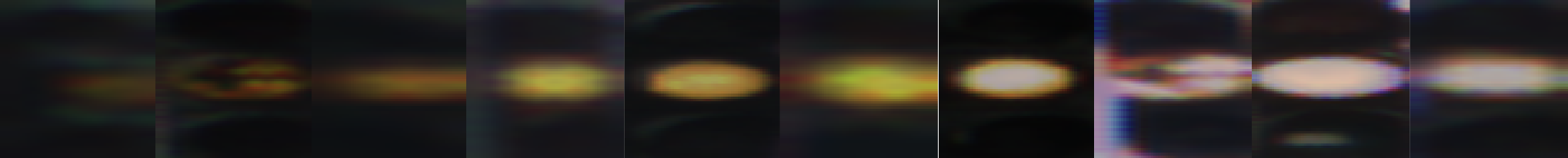}\label{fig:driveu_yellow5_traversal}} \vspace{-1em} \\
	\subfloat[$z_{20}$ controls the shade of red color light]{\includegraphics[width=0.47\textwidth]{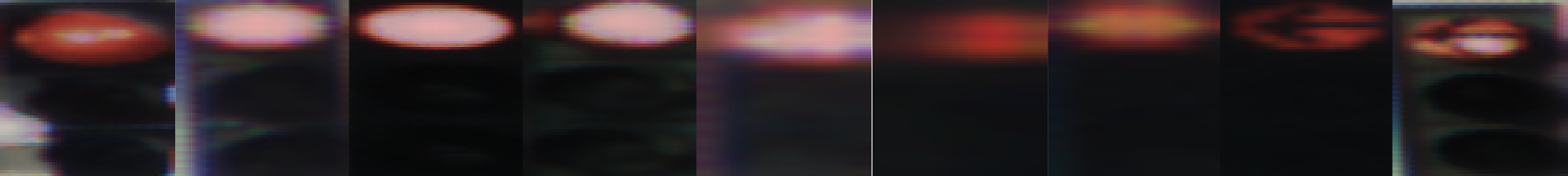}\label{fig:driveu_red_20}} \vspace{-1em} \\
	\subfloat[$z_{23}$ controls the blurriness of the red yellow light]{\includegraphics[width=0.47\textwidth]{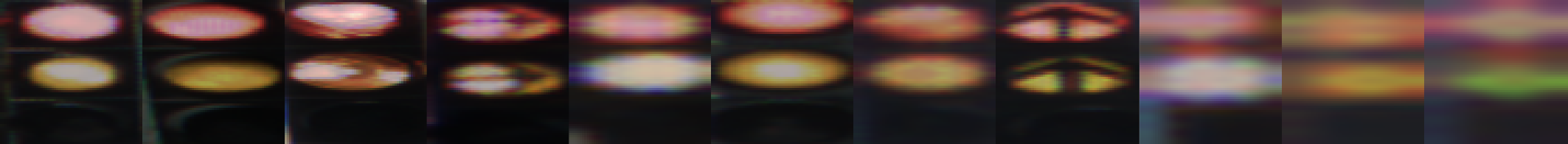}\label{fig:driveu_red_yelow_23}} 
	\caption{Traversal of VAE latent dimensions trained on DriveU dataset}
	\label{fig:rel_traversals_driveu}
	\vspace{-1em}
\end{figure}
\begin{figure}
	\vspace{-2em}
	\centering
	\subfloat{\includegraphics[width=0.48 \textwidth, keepaspectratio]{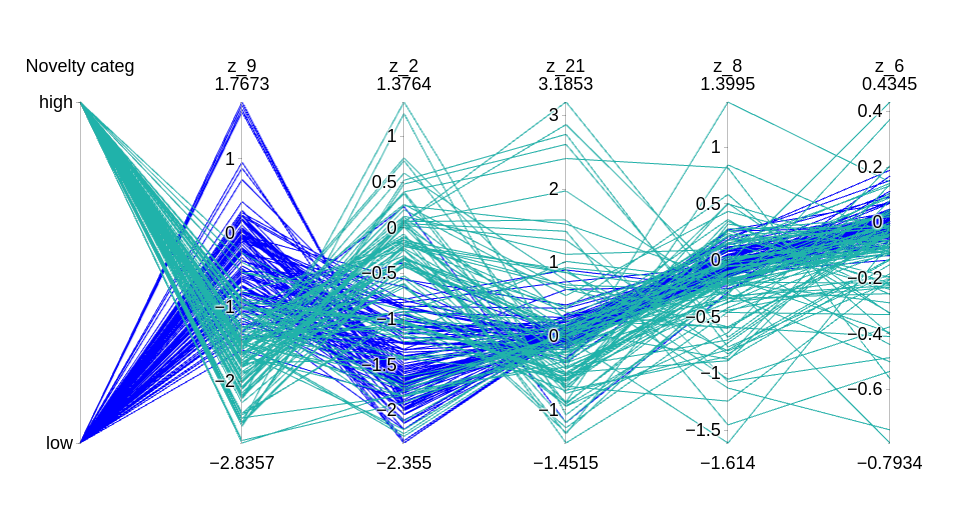}\label{fig:bstld_green_pc_kde}} \vspace{-3em} \\
	\subfloat{\includegraphics[width=0.48 \textwidth, keepaspectratio]{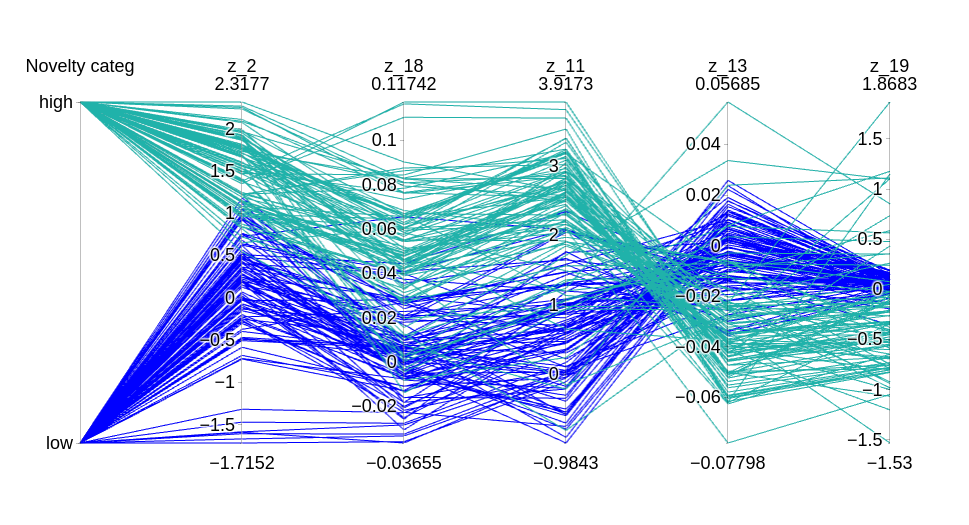}\label{fig:bstld_red_pc_kde}}
	\caption{Visualization of sample novelty based on KDE density scores for green (top) and red (bottom) traffic lights in BSTLD dataset}
	\label{fig:bstld_red_green_pc}
	\vspace{-2em}
\end{figure}
\begin{figure}
    \vspace{-1em}
	\centering
	\subfloat{\includegraphics[width=0.48 \textwidth, keepaspectratio]{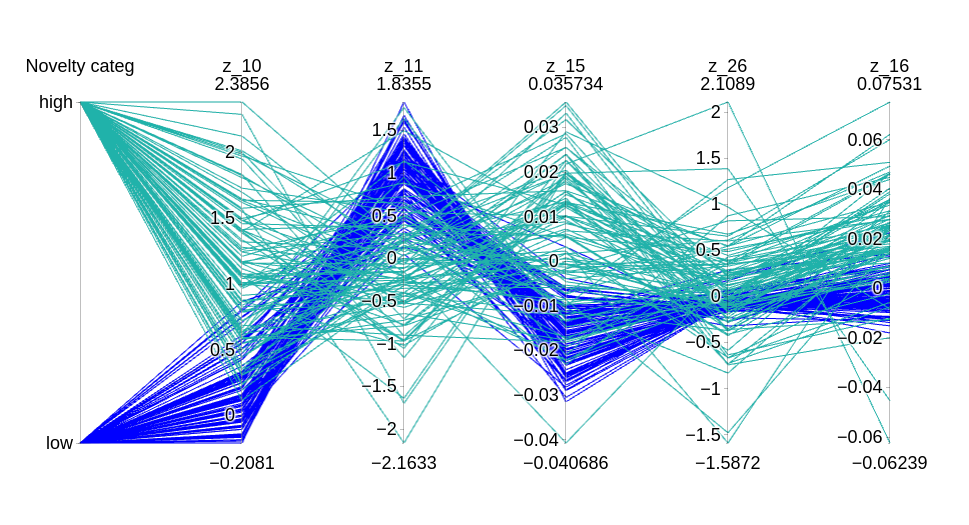}\label{fig:driveu_green_all}} \vspace{-3em} \\
	\subfloat{\includegraphics[width=0.48 \textwidth, keepaspectratio]{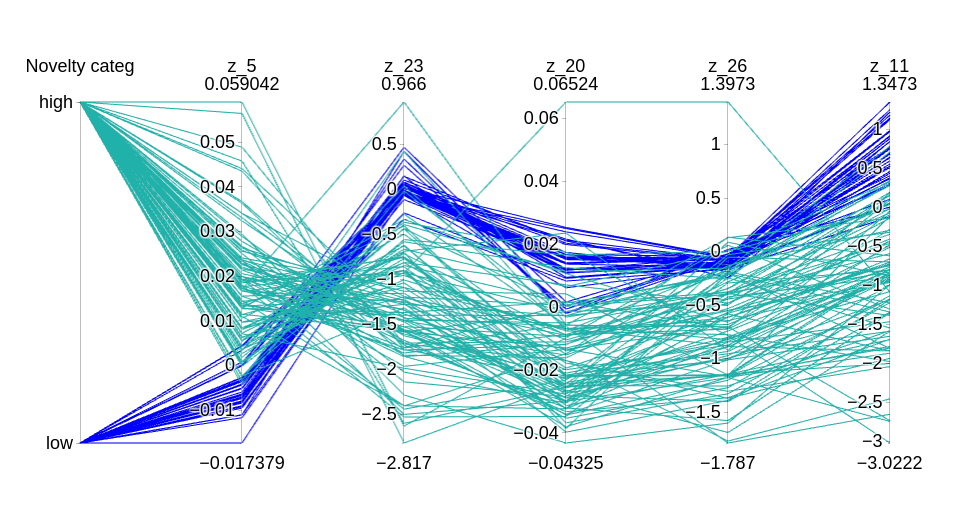}\label{fig:driveu_red_yell_all}} \vspace{-3em} \\
	\subfloat{\includegraphics[width=0.48 \textwidth, keepaspectratio]{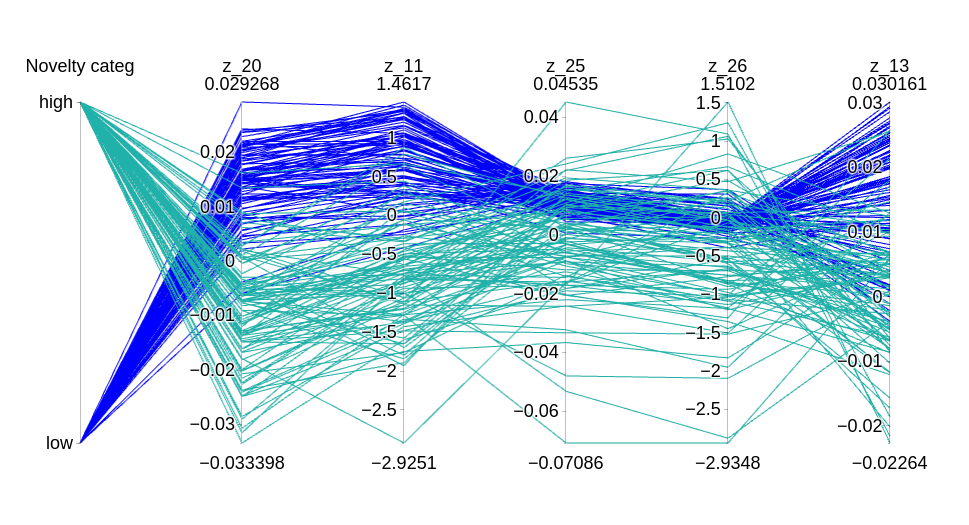}\label{fig:driveu_red_all}} \vspace{-3em} \\
	\subfloat{\includegraphics[width=0.48 \textwidth, keepaspectratio]{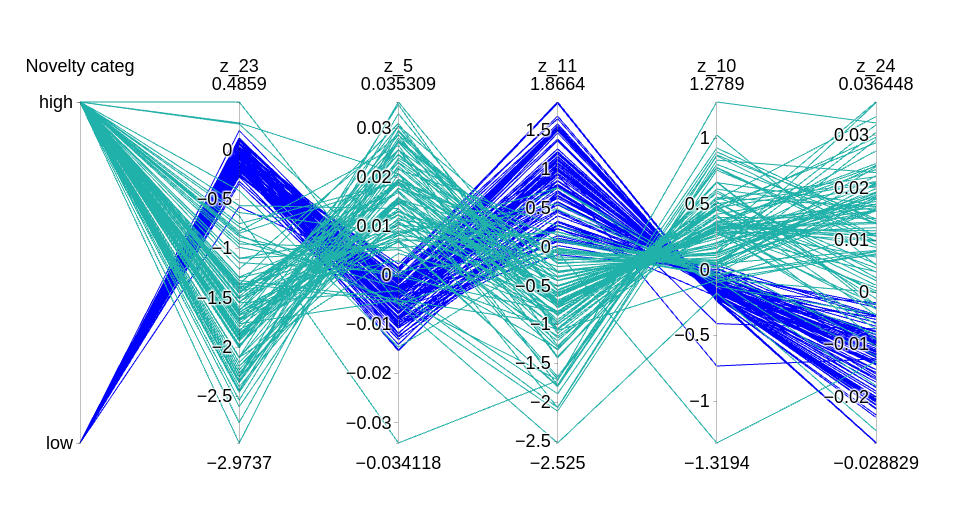}\label{fig:driveu_yellow_all}}
	\caption{Visualization of sample novelty based on KDE density scores for green, red-yellow, red and yellow traffic lights in DriveU dataset}
	\label{fig:driveu_red_green_yel_redyel_pc}
	\vspace{-2em}
\end{figure}
In Figure \ref{fig:bstld_red_green_pc} and \ref{fig:driveu_red_green_yel_redyel_pc} we map the data samples and the traversals using parallel coordinates. 
Although we trained the detector for localization and not classification, we evaluate class-wise novelty as we found them to be more straightforward. However, the scoring was done over all colors at one shot and not class-wise. 
In Figure \ref{fig:bstld_red_green_pc} we show the green and red TL samples of BSTLD dataset. 
We infer that dimension $z_9$ influences the novelty of green lights, i.e., smaller values of $z_9$ are scored as high novel samples. 
From our traversal study on $z_9$ in Figure \ref{fig:bstld_green_traversal_d9} we interpret that it controls the size of the light bulb. 
That is, the $z_9$ is inversely proportional to the bulb size or the training data has more of smaller traffic lights. 
This leads to the KDE assigning larger lights as more novel objects. 
The $z_2$ dimension is commonly influencing the novelty in green and red lights, from Figure \ref{fig:bstld_red_traversal_d2} we understand that it controls the background brightness. 
The novelty scorer assigns the TLs with bright background as more novel. 
Similarly, for red TLs in Figure \ref{fig:bstld_red_green_pc} we observe that smaller values of $z_{19}$ are more novel. 
The dimension $z_{19}$ controls the inlay, as we increase the value from left to right the image changes from a left facing arrow to red circled inlay (c.f. Figure \ref{fig:bstld_red_arrow_19}). 
From the perspective of human-interpretation, this means that the training dataset had fewer arrow inlays compared to the circle inlays. 

Likewise, the interpretation study on the DriveU dataset for four different TL colors shows that novelty increases with increase in $z_{10}$ or $z_5$ (c.f. Figure \ref{fig:driveu_red_green_yel_redyel_pc}). 
On contrary, the novelty of samples increase with the decrease of $z_{20}$ and $z_{23}$. 
From Figure \ref{fig:driveu_green_all}, we understand that $z_{10}$ controls the traffic light size and also the bright spot around it. 
These large TLs with such glaring white spots are deemed novel by the KDE. 
The $z_5$ dimension controls the shade of yellow, as we traversed to higher values we observed that the color was much darker, i.e., it tends to look more red in color, and it grew in size (c.f. Figure \ref{fig:driveu_red_yell_all}). 
The $z_{20}$ dimension regulates the shade of red light color and $z_{23}$ controls the blurriness of the lights (c.f. Figure \ref{fig:driveu_red_all} and \ref{fig:driveu_yellow_all} respectively).

%% file: conclusion.tex
\section{Conclusion and future work}\label{sec:conclusion} 
The fitness of the CNN depends on its ability to perform well on unseen data. However, existing works define unseen data as only the data not involved in the training. 
Such oversimplified definition of test data does not hold well for applications such as autonomous driving where we have streams of data from multiple sources. 
In this work we introduced the idea of scoring the generalization performance of CNN based on the sample's novelty w.r.t. the training data. 
We demonstrated its usefulness in traffic light detection application for autonomous driving domain. 
Moreover, we experimentally compared various novelty scoring algorithm that fits for our application. 
In addition to our experimental evaluations on two datasets and CNN architectures, we use visualization methods to help users interpret the notion of novelty. 

In this work, we evaluated the framework for TL applications which have standard shapes. 
However, evaluating the framework on non-standard shapes like pedestrian or vehicles is still an open challenge that we intend to pursue as future work.

%% file: main.bbl
\begin{thebibliography}{10}\itemsep=-1pt

\bibitem{angiulli2010outlier}
Fabrizio Angiulli, Rachel Ben-Eliyahu-Zohary, and Luigi Palopoli.
\newblock Outlier detection for simple default theories.
\newblock {\em Artificial Intelligence}, 174(15):1247--1253, 2010.

\bibitem{robustae}
Laura Beggel, Michael Pfeiffer, and Bernd Bischl.
\newblock Robust anomaly detection in images using adversarial autoencoders.
\newblock In {\em Machine Learning and Knowledge Discovery in Databases -
  European Conference, {ECML} {PKDD} 2019, W{\"{u}}rzburg, Germany, September
  16-20, 2019, Proceedings, Part {I}}, volume 11906, pages 206--222. Springer,
  2019.

\bibitem{bstld}
Karsten Behrendt, Libor Novak, and Rami Botros.
\newblock A deep learning approach to traffic lights: Detection, tracking, and
  classification.
\newblock In {\em Robotics and Automation (ICRA), 2017 IEEE International
  Conference on}, pages 1370--1377. IEEE, 2017.

\bibitem{lof}
Markus~M Breunig, Hans-Peter Kriegel, Raymond~T Ng, and J{\"o}rg Sander.
\newblock Lof: identifying density-based local outliers.
\newblock In {\em Proceedings of the 2000 ACM SIGMOD international conference
  on Management of data}, pages 93--104, 2000.

\bibitem{betavae}
Christopher~P Burgess, Irina Higgins, Arka Pal, Loic Matthey, Nick Watters,
  Guillaume Desjardins, and Alexander Lerchner.
\newblock Understanding disentangling in {$\beta$}-vae.
\newblock 2018.

\bibitem{noveltyocsvm}
J.~A. Carino, D. Zurita, A. Picot, M. Delgado, J.~A. Ortega, and R.~J.
  Romero-Troncoso.
\newblock Novelty detection methodology based on multi-modal one-class support
  vector machine.
\newblock In {\em 2015 IEEE 10th International Symposium on Diagnostics for
  Electrical Machines, Power Electronics and Drives (SDEMPED)}, pages 184--190,
  2015.

\bibitem{chen2018autoencoder}
Zhaomin Chen, Chai~Kiat Yeo, Bu~Sung Lee, and Chiew~Tong Lau.
\newblock Autoencoder-based network anomaly detection.
\newblock In {\em 2018 Wireless Telecommunications Symposium (WTS)}, pages
  1--5. IEEE, 2018.

\bibitem{ding2014experimental}
Xuemei Ding, Yuhua Li, Ammar Belatreche, and Liam~P Maguire.
\newblock An experimental evaluation of novelty detection methods.
\newblock {\em Neurocomputing}, 135:313--327, 2014.

\bibitem{noveltydiscreteseq}
R{\'e}mi Domingues, Pietro Michiardi, J{\'e}r{\'e}mie Barlet, and Maurizio
  Filippone.
\newblock A comparative evaluation of novelty detection algorithms for discrete
  sequences.
\newblock {\em Artificial Intelligence Review}, pages 1--26, 2019.

\bibitem{dtld}
A. Fregin, J. Müller, U. Kreβel, and K. Dietmayer.
\newblock The driveu traffic light dataset: Introduction and comparison with
  existing datasets.
\newblock In {\em 2018 IEEE International Conference on Robotics and Automation
  (ICRA)}, pages 3376--3383, May 2018.

\bibitem{general_defn}
Kevin Fritz, Daniel K{\"o}nig, Ulrich Klauck, and Michael Teutsch.
\newblock Generalization ability of region proposal networks for multispectral
  person detection.
\newblock In {\em Automatic Target Recognition XXIX}, volume 10988, page
  109880Y. International Society for Optics and Photonics, 2019.

\bibitem{goldstein2012histogram}
Markus Goldstein and Andreas Dengel.
\newblock Histogram-based outlier score (hbos): A fast unsupervised anomaly
  detection algorithm.
\newblock {\em KI-2012: Poster and Demo Track}, pages 59--63, 2012.

\bibitem{goodfellow2016deep}
Ian Goodfellow, Yoshua Bengio, Aaron Courville, and Yoshua Bengio.
\newblock {\em Deep learning}, volume~1.
\newblock MIT press Cambridge, 2016.

\bibitem{VATLD}
Liang Gou, Lincan Zou, Nanxiang Li, Michael Hofmann, Arvind~Kumar Shekar, Axel
  Wendt, and Liu Ren.
\newblock {VATLD:} {A} visual analytics system to assess, understand and
  improve traffic light detection.
\newblock {\em {IEEE} Trans. Vis. Comput. Graph.}, 27(2):261--271, 2021.

\bibitem{he2019control}
Fengxiang He, Tongliang Liu, and Dacheng Tao.
\newblock Control batch size and learning rate to generalize well: Theoretical
  and empirical evidence.
\newblock In H. Wallach, H. Larochelle, A. Beygelzimer, F. d\textquotesingle
  Alch\'{e}-Buc, E. Fox, and R. Garnett, editors, {\em Advances in Neural
  Information Processing Systems}, volume~32. Curran Associates, Inc., 2019.

\bibitem{7004298}
M. {Johanson}, S. {Belenki}, J. {Jalminger}, M. {Fant}, and M. {Gjertz}.
\newblock Big automotive data: Leveraging large volumes of data for
  knowledge-driven product development.
\newblock In {\em 2014 IEEE International Conference on Big Data (Big Data)},
  pages 736--741, 2014.

\bibitem{lee2015novelty}
Changyong Lee, Bokyoung Kang, and Juneseuk Shin.
\newblock Novelty-focused patent mapping for technology opportunity analysis.
\newblock {\em Technological Forecasting and Social Change}, 90:355--365, 2015.

\bibitem{lian2012feature}
Heng Lian.
\newblock On feature selection with principal component analysis for one-class
  svm.
\newblock {\em Pattern Recognition Letters}, 33(9):1027--1031, 2012.

\bibitem{liu2008isolation}
Fei~Tony Liu, Kai~Ming Ting, and Zhi-Hua Zhou.
\newblock Isolation forest.
\newblock In {\em 2008 eighth ieee international conference on data mining},
  pages 413--422. IEEE, 2008.

\bibitem{markou2003novelty}
Markos Markou and Sameer Singh.
\newblock Novelty detection: a review—part 1: statistical approaches.
\newblock {\em Signal processing}, 83(12):2481--2497, 2003.

\bibitem{masud2010classification}
Mohammad Masud, Jing Gao, Latifur Khan, Jiawei Han, and Bhavani~M
  Thuraisingham.
\newblock Classification and novel class detection in concept-drifting data
  streams under time constraints.
\newblock {\em IEEE Transactions on Knowledge and Data Engineering},
  23(6):859--874, 2010.

\bibitem{mcinnes2018umap}
Leland McInnes, John Healy, and James Melville.
\newblock Umap: Uniform manifold approximation and projection for dimension
  reduction.
\newblock {\em arXiv preprint arXiv:1802.03426}, 2018.

\bibitem{mendelson2020online}
Shon Mendelson and Boaz Lerner.
\newblock Online cluster drift detection for novelty detection in data streams.
\newblock In {\em 2020 19th IEEE International Conference on Machine Learning
  and Applications (ICMLA)}, pages 171--178. IEEE, 2020.

\bibitem{reviewofnovelty}
Dubravko Miljkovi{\'c}.
\newblock Review of novelty detection methods.
\newblock In {\em The 33rd International Convention MIPRO}, pages 593--598.
  IEEE, 2010.

\bibitem{pevny2016loda}
Tom{\'a}{\v{s}} Pevn{\`y}.
\newblock Loda: Lightweight on-line detector of anomalies.
\newblock {\em Machine Learning}, 102(2):275--304, 2016.

\bibitem{ocsvm}
Bernhard Sch{\"o}lkopf, Robert~C Williamson, Alexander~J Smola, John
  Shawe-Taylor, John~C Platt, et~al.
\newblock Support vector method for novelty detection.
\newblock In {\em NIPS}, volume~12, pages 582--588. Citeseer, 1999.

\bibitem{ijcvarvind}
Arvind~Kumar Shekar, Liang Gou, Liu Ren, and Axel Wendt.
\newblock Label-free robustness estimation of object detection cnns for
  autonomous driving applications.
\newblock {\em Int. J. Comput. Vis.}, 129(4):1185--1201, 2021.

\bibitem{tan2011fast}
Swee~Chuan Tan, Kai~Ming Ting, and Tony~Fei Liu.
\newblock Fast anomaly detection for streaming data.
\newblock In {\em Twenty-Second International Joint Conference on Artificial
  Intelligence}, 2011.

\bibitem{vasilev2020q}
Aleksei Vasilev, Vladimir Golkov, Marc Meissner, Ilona Lipp, Eleonora Sgarlata,
  Valentina Tomassini, Derek~K Jones, and Daniel Cremers.
\newblock q-space novelty detection with variational autoencoders.
\newblock In {\em Computational Diffusion MRI}, pages 113--124. Springer, 2020.

\bibitem{zhao2019pyod}
Yue Zhao, Zain Nasrullah, and Zheng Li.
\newblock Pyod: A python toolbox for scalable outlier detection.
\newblock {\em arXiv preprint arXiv:1901.01588}, 2019.

\bibitem{zheng2018improvement}
Qinghe Zheng, Mingqiang Yang, Jiajie Yang, Qingrui Zhang, and Xinxin Zhang.
\newblock Improvement of generalization ability of deep cnn via implicit
  regularization in two-stage training process.
\newblock {\em IEEE Access}, 6:15844--15869, 2018.

\end{thebibliography}
